\documentclass{ecai}

\usepackage{latexsym}
\usepackage{amssymb}
\usepackage{amsmath}
\usepackage{amsthm}
\usepackage{booktabs}
\usepackage{enumitem}
\usepackage{graphicx}
\usepackage{color}
\usepackage{graphicx}%
\usepackage{multirow}%
\usepackage{amsmath,amssymb,amsfonts}%
\usepackage{amsthm}%
\usepackage{mathrsfs}%
\usepackage[title]{appendix}%
\usepackage{xcolor}%
\usepackage{textcomp}%
\usepackage{manyfoot}%
\usepackage{booktabs}%
\usepackage{algorithm}%
\usepackage{algorithmicx}%
\usepackage{algpseudocode}%
\usepackage{listings}%
\usepackage{bm}%
\usepackage{todonotes}%
\usepackage[utf8]{inputenc}
\usepackage{mathtools}
\usepackage{url}

\usepackage{dsfont}

\theoremstyle{thmstyleone}%
\newtheorem{theorem}{Theorem}[section]

\newtheorem{lemma}[theorem]{Lemma}

\theoremstyle{thmstyletwo}%
\newtheorem{assumption}{Assumption}

\def\app#1#2{%
  \mathrel{%
    \setbox0=\hbox{$#1\sim$}%
    \setbox2=\hbox{%
      \rlap{\hbox{$#1\propto$}}%
      \lower1.1\ht0\box0%
    }%
    \raise0.25\ht2\box2%
  }%
}

\newcommand{\appropto}{\mathrel{\vcenter{
  \offinterlineskip\halign{\hfil$##$\cr
    \propto\cr\noalign{\kern2pt}\sim\cr\noalign{\kern-2pt}}}}}

\newtheorem{goal}{Aim}

\theoremstyle{thmstylethree}%
\newtheorem{definition}{Definition}%

\usepackage{amsthm}
\newtheorem{innercustomgeneric}{\customgenericname}
\providecommand{\customgenericname}{}
\newcommand{\newcustomtheorem}[2]{%
  \newenvironment{#1}[1]
  {%
   \renewcommand\customgenericname{#2}%
   \renewcommand\theinnercustomgeneric{##1}%
   \innercustomgeneric
  }
  {\endinnercustomgeneric}
}

\newcustomtheorem{customthm}{Theorem}
\newcustomtheorem{customlemma}{Lemma}
\newcustomtheorem{customprop}{Proposition}

\newcommand{\BibTeX}{B\kern-.05em{\sc i\kern-.025em b}\kern-.08em\TeX}

\newcommand{\sign}{\operatorname{sign}}

\begin{document}

\begin{frontmatter}

\title{GRADSTOP: Early Stopping of Gradient Descent via Posterior Sampling}

\author[A]{\fnms{Arash} \snm{Jamshidi}\thanks{Corresponding Author. Email: arash.jamshidi@helsinki.fi.}}%

\author[A]{\fnms{Lauri} \snm{Sepp\"al\"ainen}} %

\author[A,B]{\fnms{Katsiaryna} \snm{Haitsiukevich}}%

\author[A]{\fnms{Hoang Phuc Hau} \snm{Luu}}%

\author[C]{\fnms{Anton} \snm{Bj\"orklund}}%

\author[A]{\fnms{Kai} \snm{Puolam\"aki}}%

\address[A]{University of Helsinki, Helsinki, Finland}

\address[B]{Aalto University, Espoo, Finland}

\address[C]{University of Oxford, Oxford, United Kingdom}

\begin{abstract}
Machine learning models are often learned by minimising a loss function on the training data using a gradient descent algorithm. These models
often suffer from overfitting, leading to a decline in predictive performance on unseen data. A standard solution is early stopping using a hold-out validation set, which halts the minimisation when the validation loss stops decreasing. However, this hold-out set reduces the data available for training. 
This paper presents {\sc gradstop}, a novel stochastic early stopping method that only uses information in the gradients, which are produced by the gradient descent algorithm ``for free.''
Our main contributions are that we estimate the Bayesian posterior by the gradient information, define the early stopping problem as drawing sample from this posterior, and use the approximated posterior to obtain a stopping criterion.
Our empirical evaluation shows that {\sc gradstop} achieves a small loss on test data and compares favourably to a validation-set-based stopping criterion.
By leveraging the entire dataset for training, our method is particularly advantageous in data-limited settings, such as transfer learning. It can be incorporated as an optional feature in gradient descent libraries with only a small computational overhead. The source code is available at \url{https://github.com/edahelsinki/gradstop}.
\end{abstract}

\end{frontmatter}

\section{Introduction}\label{sec1}

Gradient descent and its variants are central tools in modern machine learning and deep learning \cite{ruder2016overview}.
Despite their simplicity, these methods are effective even for large neural networks with billions of parameters trained on large-scale datasets \cite{achiam2023gpt,abramson2024accurate}.
However, if the training data is limited or the model is not sufficiently regularised, a long optimisation process tends to result in overfitting and poor generalisation properties of the model.
This motivates the practical use of various regularisation techniques \cite{sietsma1991creating,goodfellow2016deep,srivastava2014dropout,ioffe2015batch}, including early stopping \cite{morgan1989generalization,zhang2005boosting}.
Early stopping is an approach that provides a stopping time for an iterative procedure such as gradient-based optimisation \cite{morgan1989generalization}.
The method is actively utilised both in the context of parametric and non-parametric models \cite{bauer2007regularization, yao2007early, raskutti2014early,wei2017early,ali2019continuous,vaskevicius2020statistical,shen2022optimal}. In validation-set-based methods, the iteration stops when the validation loss stops decreasing or increases.

The validation-set-based early stopping is reasonable when there is enough data to reserve a sufficiently large validation set without hurting the model's performance. Such large data sets are typical in many deep-learning setups but may not be available in transfer learning scenarios or for applications with costly data collection.
Thereby, using all the available data for training is desirable if the data is scarce. There is a clear need for a simple, efficient, and theoretically justified early stopping criterion. Since most machine learning models rely on gradient-based methods, we design our stopping criterion to be compatible with off-the-shelf gradient descent. Addressing this need, the practical aim of this paper is as follows:
\begin{goal}\label{goal:practical} [practical]
    Provide a theoretically justified early stopping method that can (i) use all data for training, (ii) does not need a validation set, and (iii) uses only the information provided by a gradient descent algorithm with a small computational overhead.
\end{goal}

In this paper, we take the Bayesian approach, where the principled way to find a representative parameter value is to draw one from the posterior distribution \cite{bernardo2000Bayesian}. In particular, we stop the gradient descent based on a draw from the posterior distribution or, alternatively, use an approximated posterior for uncertainty estimation. In principle, if we had access to the exact posterior distribution, we could draw samples directly, e.g., by using Markov chain Monte-Carlo methods (MCMC) \cite{gelman2014Bayesian}. However, MCMC is a computationally much more expensive and time-consuming procedure that can be considered a Bayesian alternative to gradient-based optimisation.
 For these reasons, we restrict posterior sampling to the sequence of model parameters produced by the gradient descent algorithm. Our method, coined as {\sc gradstop}, chooses the stopping point for which the distribution of the posterior probabilities would most closely match a draw from the posterior.  

We assume that the $n$ training data points $D=(z_1,\ldots,z_n)$ have been drawn independently from a fixed but unknown distribution $p(z\mid\theta)$, where $\theta\in{\mathbb{R}}^d$ denotes the model parameters. 
We aim to find the parameters $\theta$ which minimise the value of a loss function
\begin{equation}\label{eq:losssum}
L(\theta)=\sum\nolimits_{i=1}^n{l_i(\theta)}
\end{equation}
expressed as a sum over terms $l_i(\theta) = l(z_i, \theta)$ that depend on the training data points. 
The underlying assumption also in the validation set methods is that the data has been drawn independently from a fixed but unknown distribution, the idea being that a loss on a validation set is a good (e.g., unbiased) estimate of the loss on new data unseen in the training \cite{hastie2009}. 

Without using a validation set, we can connect the distribution that generated the data to the loss function by assuming the loss function is the negative log-posterior, or $L(\theta)=-\log{p(\theta\mid D)}+{\rm const}$, with the posterior $p(\theta\mid D)\propto e^{-L(\theta)}$. This is satisfied if we choose
\begin{equation}\label{eq:li}
l_i(\theta)=l(z_i,\theta)=-\log{p(z_i\mid\theta)}-\frac 1n\log{p(\theta)},
\end{equation}
 similar to Eqs. (2)--(3) of \cite{pmlr-v48-mandt16}.  Note that in our probabilistic framing, $p(\theta)$ is the prior distribution and $p(z_i \mid \theta)$ is the likelihood.
 Often, the loss function of a machine learning model can be written in the form of Eq.~\eqref{eq:losssum}, even if the model is not probabilistic; in this case, we can define the corresponding likelihood function and prior term by matching the loss with Eq. \eqref{eq:li}. We show in Sect. \ref{sec:results} that our approach also works for non-probabilistic models, as long as the loss function is reasonably behaving and of the form given by Eq. \eqref{eq:losssum}.

A gradient-based optimisation algorithm produces a sequence of $T$ parameter values $\Theta = \{\theta_1,\ldots,\theta_T\}$, converging to a local minimum of the loss function, and the respective gradient vectors,
\begin{equation}\label{eq:gi}
g_i(\theta)=g(z_i,\theta)=\nabla_\theta{l_i(\theta)}.
\end{equation}
By accessing only $\theta_t$ and $g_i(\theta_t)$, our objective is to draw $t\in[T]=\{1,\ldots,T\}$ such that $\theta_t$ obeys ``as closely as possible'' the posterior distribution $p(\theta\mid D)$. 
From here, we define the computational equivalent to the gradient stopping in Aim~\ref{goal:practical}.
\begin{goal}\label{goal:main}[computational]
Given the definitions above and using the sequence of parameters $\theta_1,\ldots,\theta_T$ and the respective gradient vectors (Eq. \eqref{eq:gi}) $g_i(\theta_t)$ for $i\in[n]$ and $t\in[T]$, draw a sample from distribution $P_T$ over $[T]$, parametrised by a vector in simplex $\Delta_T=\{P_T\in{\mathbb{R}}_{\ge 0}^T\mid\sum\nolimits_{t=1}^T{P_T(t)}=1\}$, 
such that the Kolmogorov distance \cite{gaunt2023bounding} between distribution of losses defined by $L(\theta)$ when $\theta=\theta_{t}$ with $t\sim P_T$, and when $\theta\sim p(\theta\mid D)$, respectively, is minimised.
\end{goal}

In summary, our contributions are as follows: (i) we define the early stopping problem as posterior sampling, (ii) we prove that our method approximates drawing a sample from the posterior distribution of the parameters of a machine learning model, and (iii) we empirically show that {\sc gradstop} achieves low test loss and is competitive with both validation-set-based stopping and early stopping methods without a separate validation set.
We evaluate the early stopping methods in data-limited settings, such as classification of medical data, transfer learning, and transfer learning with noisy labels. 

The rest of the paper is organised as follows: Sect.~\ref{sec:overview} provides the intuition behind our method and describes its implementation for practical use, while Sect.~\ref{sec:related} covers related work. 
Sect.~\ref{sec:theory} presents
the theoretical foundation of {\sc gradstop}, which utilises only the gradients and the covariance matrix of gradients to approximate sampling from the posterior distribution. We follow the theoretical section with practical considerations in Sect.~\ref{sec:practical_considerations}.
Sect.~\ref{sec:results} provides an empirical comparison of the method with the baselines.
Sect.~\ref{sec:future_work} and Sect.~\ref{sec:conclusion} discuss the directions for future work and conclude the paper.

\begin{figure*}
    \centering
    \includegraphics[width=0.45\linewidth,height = 5.8cm]{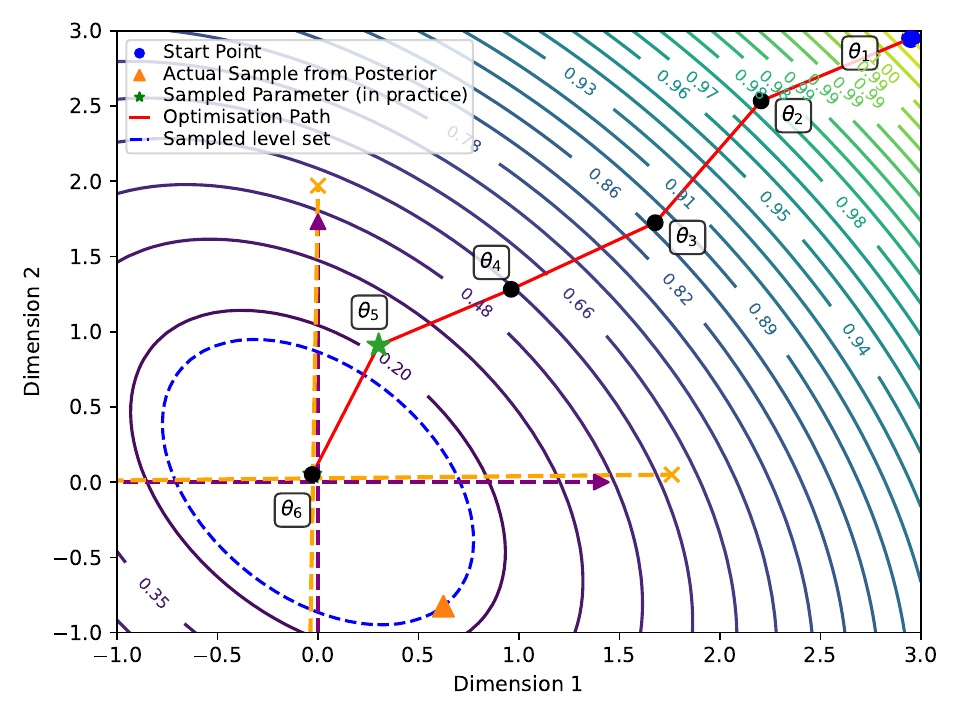}
    \hspace{0em}
    \raisebox{-0.5mm}{ %
        \includegraphics[width=0.46\linewidth,height = 5.82cm]{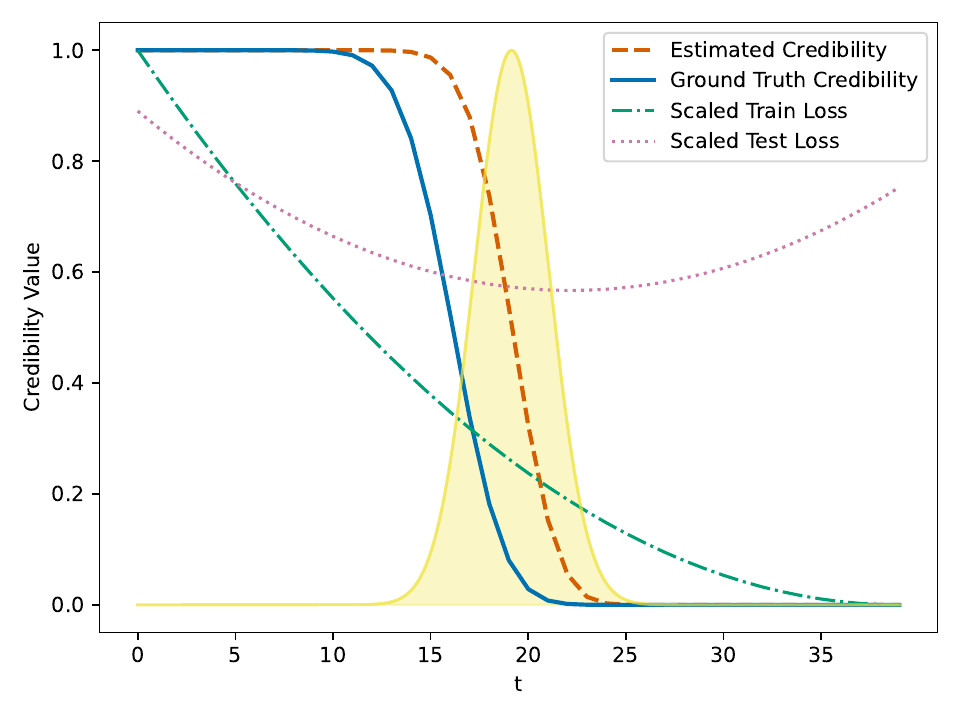}
    }
    \vspace{-0.4cm}
    \caption{\textbf{{\sc gradstop}: sampling from the posterior restricted to parameter values as a stopping criterion.} 
The left image illustrates the idea behind our method, which is to approximate sampling from the posterior by first sampling a level set (a posterior contour, dashed blue line), and then choosing the point $\theta_t$ (green star) from the set of parameters ($\theta_1, \dots, \theta_6$) produced by the optimisation algorithm that is closest to the sampled level set.
 The orange triangle is a sample from the full posterior (not available in practice). The purple arrows show the approximated standard deviation for the two dimensions and the orange cross shows the true standard deviations. The right image shows that in minimising high-dimensional quadratic losses, our method estimates the credibility values (Eq. \eqref{eq:cred}) with good accuracy along the gradient descent path (blue and orange lines). The distribution of sampled stopping times $t$ is shown in yellow.}
    
    \label{fig:sample_project}
\end{figure*}

\section{Overview of the {\sc gradstop} Algorithm}\label{sec:overview}

In this section we describe the {\sc gradstop} algorithm and the high-level intuition behind it. Its properties are formally derived in Sect.~\ref{sec:theory}.

First, we define the credibility value, which is the statistic used in the proposed stopping criterion for gradient-based optimisation.
\begin{definition}\label{def:pvalue}
    For any $\theta \in \mathbb{R}^d$ we define its credibility value as
    \begin{equation}\label{eq:cred}
    s(\theta\mid D) 
=%
\int\nolimits_{\theta': p(\theta' \mid D) \leq p(\theta \mid D)}{p(\theta' \mid D)} d \theta'
\end{equation}
which is the probability mass of the posterior within the sublevel set of the posterior at level $p(\theta \mid D)$.
\end{definition}

\begin{algorithm}[h]
\caption{The {\sc gradstop} algorithm.\label{alg:gradstop}}
\begin{algorithmic}[1]
\Require An optimisation algorithm $O$, parametrised by $t\in[T]$, that outputs a parameter value $\theta_t$ and the corresponding gradients ${\bf G}_t$ for the $t$-th gradient step. ${\bf G}_t\in{\mathbb{R}}^{n\times d}$ denotes the matrix of gradient vectors where the $i$-th row of is given by $g_i(\theta_t)$; see Eq. \eqref{eq:gi}, calculated by backprop.
\Ensure The parameter value that best matches drawing a parameter from the posterior.
\State $u\sim U(0,1)$ \Comment{Draw a random variable from the uniform distribution or use a preset constant value; see Sec. \ref{sec:practical_considerations}.}
\State $s_{best}\leftarrow \infty$
\For{$t = 1$ to $T$}
\State $\theta_t,{\bf G}_t\leftarrow O(t)$
\State $s_t\leftarrow \hat s({\bf G}_t)$\Comment{see Alg. \ref{alg:credible} and Eq. \eqref{eq:shat}.}
\If{$\left | s_t - u \right|< \left | s_{best} -u \right|$}
\State $\theta_{best}\leftarrow \theta_t$
\State $s_{best}\leftarrow s_t$
\EndIf
\EndFor
\State {\bf return} $\theta_{\text{best}}$.
\end{algorithmic}
\end{algorithm}

\begin{algorithm}[h]
\caption{The {\sc crediblevalue} algorithm.\label{alg:credible}}
\begin{algorithmic}[1]
\Require A matrix of gradient vectors, ${\bf G}\in{\mathbb{R}}^{n\times d}$, where the $i$-th row of is given by $g_i(\theta)$, see Alg. \ref{alg:gradstop}. 
\Ensure $\hat s({\bf G})$, the credible value.
\State ${\overline{\bf g}}\leftarrow {\bf e}_n^\intercal{\bf G}/n$\Comment{Average gradient, ${\bf e}_n=(1,\ldots,1)^\intercal\in{\mathbb{R}}^n$.}
\State $\overline{\bf G}\leftarrow({\bf G}-{\bf e}_n\overline{\bf g}^\intercal)/\sqrt{n}$\Comment{Centred gradients.}
\If{$d\le n$}\Comment{Efficient matrix inversion; see Sect. \ref{sec:practical_considerations}.}
\State $\Sigma_G\leftarrow \overline{\bf G}^\intercal\overline{\bf G}$ \Comment{Estimated covariance matrix; see Eq. \eqref{def:covmatrix}.}
\State $\hat \Sigma_G\leftarrow(1-\epsilon)\Sigma_G+\epsilon\,{\rm tr}(\Sigma_G){\bf 1}_d/d$ %
\\ \Comment{OAS covariance estimate; see Sect. \ref{sec:practical_considerations}}
\State ${\bf A}\leftarrow\hat \Sigma_G^{-1}$ \Comment{Standard matrix inversion.}
\Else{}
\State $\delta \leftarrow (\epsilon{\rm tr}(\overline{\bf G}^\intercal\overline{\bf G})/d)^{-1} $
\State ${\bf A}\leftarrow\delta{\bf 1}_d-\delta(1-\epsilon)\overline{\bf G}^\intercal\left(\delta{\bf 1}_d+(1-\epsilon)\overline{\bf G}~\overline{\bf G}^\intercal\right)^{-1}\overline{\bf G}$ \\ \Comment{Matrix inversion using the Woodbury matrix identity.}
\EndIf
\State $z\leftarrow n\overline{\bf g}^\intercal{\bf A}\overline{\bf g}$ \Comment{$z$ can be computed by a least squares solver.}
\State {\bf return} $1-F_{\chi^2_d}(z)$.\Comment{$\hat s({\bf G})$; see Thm. \ref{thm:main}.}
\end{algorithmic}
\end{algorithm}

The target of a gradient descent optimisation algorithm $O$ is to minimise the loss $L(\theta)$ given by Eq. \eqref{eq:losssum}, which is a sum of $n$ terms given by Eq. \eqref{eq:li}. Each term is specific to a training data point, and is proportional to the negative logarithm of the posterior---or log-loss plus regularisation terms. The algorithm $O$ outputs a sequence of parameter values $\theta_1,\ldots,\theta_T$, shown schematically in Fig. \ref{fig:sample_project}, converging towards the (local) minimum of the loss (or, equivalently, maximum-a-posteriori solution). 
{\sc gradstop} selects the value of $\theta_t$ that is the closest to the sampled %
equal-value contour from the posterior by its ``credibility value'' (probability mass within the contour).

In practice (see Alg. \ref{alg:gradstop}), we first draw a random number $u$ from a uniform distribution in $[0,1]$ or use a preset constant value (line 1). Next, we run the optimisation algorithm (lines 3---10) and find the parameter value $\theta_t$, whose credibility value is closest to $u$. The gradient descent algorithm $O$ outputs the parameter values $\theta_t$ and the respective gradients (line 4). As we show later in Sect. \ref{sec:theory}, we can use this gradient information to estimate the credibility value (line 5 and Alg. \ref{alg:credible}) and to select the best $\theta_t$. In the implementation, we only need to store one extra copy of the parameter values ($\theta_{best}$ in line 7).

\begin{figure}[t]
    \centering
    \includegraphics[width=\linewidth,trim={0 2 0 10},clip]{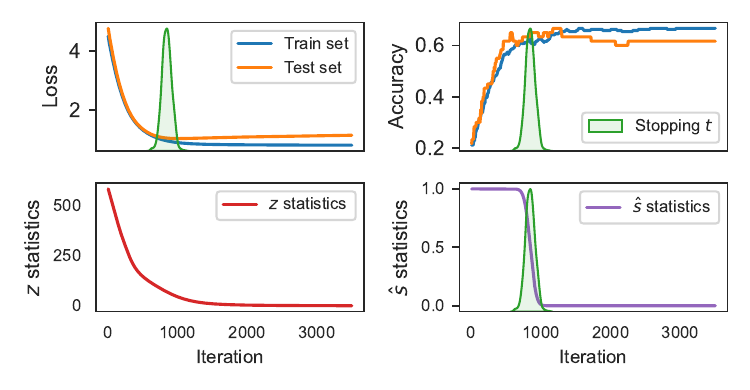}%
    \vspace{-0.9cm}
    \caption{\textbf{Demonstration of {\sc gradstop} statistics for classification of heart diseases with logistic regression.} The figures show training (blue) and test (orange) loss values, accuracies, and statistics $z$ (Alg.~\ref{alg:credible}, line 11) and $\hat{s}$ (Eq. \eqref{eq:shat}) through training. The distribution of sampled stopping times $t$ is in green. {\sc gradstop} finds an iteration of the optimisation process before the test loss increases preventing overfitting. It relies on values of $\hat{s}$ statistics calculated with training data only.}
    \label{fig:linear_stop}
\end{figure}

The credibility values are calculated according to Alg.~\ref{alg:credible} from the matrix of gradient vectors. As an example, we demonstrate intermediate steps in the calculation during the optimisation process of logistic regression for a classification task of heart diseases in Fig.~\ref{fig:linear_stop}.

Since the credible values follow a uniform distribution on $[0,1]$ (as $u$ does), our algorithm—when it accurately estimates these credible values—outputs a parameter value whose negative log‐posterior (or, equivalently, posterior probability) distribution closely matches that of a parameter sampled from the true posterior, as required by Aim~\ref{goal:main}. In other words, applying inverse of CDF (or inverse of 1 – CDF) to a uniformly distributed random variable u on [0,1] yields a sample from the original distribution \cite{ross2013first}.

\section{Related Work}\label{sec:related}

Early stopping can be considered a form of regularisation \cite{bishop2006pattern,sjoberg1995overtraining}. In the case of a simple linear model with quadratic loss, early stopping is equivalent to $\ell_2$ regularisation \cite{friedman2004gradient,goodfellow2016deep}. Under certain assumptions, early stopping also improves model robustness to noise in additional ways compared to the standard $\ell_2$ regularisation. 
The authors of \cite{li2020gradient} demonstrate that gradient descent with early stopping is provably robust to corruption in a constant fraction of the labels. Similarly, works \cite{nguyen2019self,bai2021understanding} empirically show the benefits of early stopping for learning in the presence of noise. 
Building on both theoretical findings and empirical evidence supporting early stopping, the current work extends its applicability by introducing novel theoretical results and a corresponding algorithm grounded in Bayesian principles.

Early stopping methods rely on a criterion to finish the optimisation. A typical criterion is tied to the model's performance on a reserved subset of training data referred to as a validation set \cite{xu2018splitting}. These methods empirically estimate the true generalisation loss by model performance on the validation set to determine the stopping point of the optimisation process \cite{morgan1989generalization,prechelt2002early,reed1993pruning,vaskevicius2020statistical,toner2024noisy}. 
While effective in practical scenarios, the methods based on validation set performance require for the validation set to be a representative sample from the true data distribution, which may not be the case. Alternatively, $k$-fold cross-validation methods can be applied to recycle the training data for validation purposes \cite{bishop2006pattern, hastie2009}, including for early stopping \cite{patil2024failures}. 
However, these methods increase the computational requirements by the number of folds (e.g., $k=8$ \cite{erickson2020autogluon}).
Furthermore, if leave-one-out cross-validation is required, as in \cite{rad2020scalable,patil2024failures}, this multiplicative factor increases to the number of samples in the dataset. Some works incorporate early stopping to cross-validation to mitigate the computational overhead \cite{bergman2024Dont}. 
In contrast to this line of work, the proposed method relies on the statistics calculated using gradients for the training set. As a result, our method does not require a separate hold-out validation set or additional model retraining.

Similar to our work, several methods \cite{raskutti2014early,mahsereci2017early,liu2020understanding,bonet2021channel,yuan2024early,fort2019stiffness,vardasbi2022intersection,forouzesh2021disparity} have data-dependent stopping criteria without a validation set. However, 
\cite{raskutti2014early} focuses on non-parametric regression with least squares loss function while our method assumes negative log posterior as a loss.
Works \cite{mahsereci2017early,bonet2021channel,yuan2024early} can be applied to a broader class of models, but they lack theoretical justifications for the proposed criteria.
Instead of utilising the gradient data directly as in our method, Vardasbi et al. \cite{vardasbi2022intersection} proposed an early stopping criterion which relies on the parameters from two initializations of the same model trained in parallel to converge. The parallel training of several model instances introduces considerable computational and memory overheads. Closest to our method are stopping criteria \cite{mahsereci2017early,forouzesh2021disparity} that utilise gradient information for the training set. Unlike our method, both works take into account only the variance of the gradients, disregarding the more complex covariance structure of the gradients.

\section{Theory}\label{sec:theory}

To show how {\sc gradstop} estimates the posterior distribution, we begin by stating our key assumption, namely that we can estimate the posterior near the MAP solution by a Gaussian distribution:
 \begin{assumption}\label{ass:gauss}
    Near the local maximum $\theta^*$ of the posterior $p(\theta\mid D)\propto e^{-L(\theta)}$, the loss can be well approximated by  $L(\theta) = (\theta-\theta^*)^\intercal H(\theta-\theta^*)/2$ with $H$ symmetric and positive definite.
\end{assumption}

Next, recall that $L(\theta)=\sum\nolimits_{i=1}^n{l_i(\theta)}$, where $l_i(\theta)=l(z_i,\theta)$. Also recall that we can express the datapoint specific loss $l(z,\theta)=-\log{p(z\mid\theta)}-\frac 1n\log{p(\theta)}$ in terms of the likelihood and the prior.  We further define an auxiliary likelihood by $q(z\mid\theta)\propto e^{-l(z\mid\theta)}$ for ease of notation in the proofs, streamlining the use of information equality (Lem. \ref{lem:infeq0}).

We denote the score vector by $g(z,\theta)=\nabla_\theta{l(z,\theta)}$ with $g_i(\theta)=g(z_i,\theta)$.
We also define the empirical gradient covariance matrix by 
\begin{align}\label{def:covmatrix}
\Sigma_G(\theta)=\sum\nolimits_{i=1}^n{(g_i(\theta)-\overline g(\theta))(g_i(\theta)-\overline g(\theta))^\intercal/n},
\end{align}
where  the mean gradient is $\overline g(\theta)=\sum\nolimits_{i=1}^n{g_i(\theta)/n}$.

Our main objective is to estimate the credibility value (Def. \ref{def:pvalue}) by information gleaned from the gradients only. We first show how to compute the credibility value from the posterior:
\begin{lemma}[Credibility Value]\label{lem:credible}
The credibility value can be written as $s(\theta\mid D)=1-F_{\chi^2_d}((\theta-\theta^*)^\intercal H(\theta-\theta^*))$, where $F_{\chi^2_d}$ is the cumulative chi-squared distribution with $d$ degrees of freedom.
\end{lemma}
\begin{proof}
   Suppose $\hat{\theta} \sim p(\theta\ \mid D) = \mathcal{N}(\theta^*, H^{-1})$.
    Hence $(\hat{\theta} - \theta^*) \sim \mathcal{N}(0,H^{-1}) \Leftrightarrow (\hat{\theta} - \theta^*) \sim H^{-\frac{1}{2}}X$ where $X \sim \mathcal{N}(0,I_d)$. Now note that we have $(\hat{\theta}-\theta^*)^\intercal H(\hat{\theta}-\theta^*) = X^\intercal H^{-\frac{1}{2}}H H^{-\frac{1}{2}}X= X^\intercal X$ and we know $X^\intercal X \sim \chi^2_d$ by the definition of the chi-squared distribution. The result follows from the definition of $s(\theta \mid D) = p_{\hat{\theta} \sim p(\theta \mid D)} \left( \mathds{1} \{p(\hat{\theta} \mid D) \leq p(\theta \mid D)\} \right)$, the pdf of $d$-dimensional normal distribution, and some algebraic manipulations.
\end{proof}

Here, we show the relation between the gradient covariance and the Hessian. We will first present two auxiliary lemmas.

\begin{lemma}[Information Equality]\label{lem:infeq0}\cite{bartlett1953Approximate}
 \begin{align}&E_{z\sim q(z\mid\theta)}\left[\nabla_\theta{\log{q(z\mid\theta)}}\nabla_\theta{\log{q(z\mid\theta)}}^\intercal\right]\nonumber\\=&-
E_{z\sim q(z\mid\theta)}\left[\nabla_\theta^2{\log{q(z\mid\theta)}}\right]\end{align} is satisfied by any distribution $q(z\mid\theta)$ that is doubly differentiable, for which differentiation and integration commute, and that is non-zero everywhere.
\end{lemma}

We use the information equality of Lem. \ref{lem:infeq0} to connect the Hessian and the gradient information. We first prove an auxiliary result.
\begin{lemma}[Hessian and Gradient]\label{lem:infeq2} 
For a fixed $\theta$ and $q(z\mid\theta)\propto e^{-l(z,\theta)}$, $E_{z\sim q(z\mid\theta)}\left[g(z,\theta)g(z,\theta)^\intercal\right]=E_{z\sim q(z\mid\theta)}\left[h(z,\theta)\right]$, where $g(z,\theta)=\nabla_\theta{l(z,\theta)}$ and $h(z,\theta)=\nabla_\theta^2{l(z,\theta)}$. 
\end{lemma}
\begin{proof}
We write the score vector by $g(z,\theta)=\nabla_\theta{l(z,\theta)}=-\nabla_\theta{\log{q(z\mid\theta)}}$ and the Hessian of the per-observation loss by $h(z,\theta)=\nabla_\theta^2{l(z,\theta)}=-\nabla_\theta^2{\log{q(z\mid\theta)}}$. 
The claim follows from the information equality (Lem. \ref{lem:infeq0}).
\end{proof}

We then show the relation between the Hessian and the gradient covariance. The following Lemma implies that $\Sigma_G\approx H/n$ holds to a reasonable accuracy around $\theta^*$.
\begin{lemma}[$\Sigma_G\approx H/n$]\label{lem:infeq3}
Asymptotically, as $n\to\infty$, assuming that $q(z\mid \theta^*)$ converges to the data generating distribution, the Hessian satisfies
$\Sigma_G(\theta^*)\xrightarrow{a.s.}  H/n$ and for any $\theta$ in a small neighborhood of $\theta^*$ we have $\Sigma_G(\theta) \approx  H/n$.

\end{lemma}

\begin{proof}
By the strong law of large numbers, we have that
$H(\theta)/n = \sum\nolimits_{i=1}^n{h(z_i,\theta)/n }\xrightarrow{a.s.} E\left[\,h(z,\theta)\right]$.
Similarly, by the law of large numbers and the continuous mapping theorem, $
\bar{g}(\theta) \xrightarrow{a.s.} E\left[g(z,\theta)\right]$,
and
$\Sigma_G(\theta) \xrightarrow{a.s.} E\left[g(z,\theta)g(z,\theta)^\intercal\right] - E\left[g(z,\theta)\right]E\left[g(z,\theta)\right]^\intercal$.
In particular, at the optimum $\theta^*$ we have $E\bigl[g(z,\theta^*)\bigr]=0$, so that
$\Sigma_G(\theta^*) \xrightarrow{a.s.} E\left[g(z,\theta^*)g(z,\theta^*)^\top\right]$.

It follows from Lem. \ref{lem:infeq2} that
$
\,\Sigma_G(\theta^*)\xrightarrow{a.s.}  H/n$.
By continuity, for all $\theta$ in a neighborhood of $\theta^*$; it holds that
$
\,\Sigma_G(\theta)\approx  H/n$.
\end{proof}

We then show the relation of the parameter values, Hessian, and the mean of the gradient.
\begin{lemma}[Mean Gradient and Hessian]\label{lem:inv}
    We have $\theta-\theta^*=n\,H^{-1}\overline g(\theta)$.
\end{lemma}
\begin{proof}
    The claim follows directly from the fact that we can write the gradient as $n\,\overline g(\theta)=\nabla_\theta{L(\theta)}=\nabla_\theta{\left((\theta-\theta^*)^\intercal H(\theta-\theta^*)/2\right)}=H(\theta-\theta^*)$.
\end{proof}

Finally, we can estimate the credible value by using the statistics computed from the gradient only.
\begin{theorem}[Credibility Value Approximation]\label{thm:main}
    The credible value can be estimated by
    \begin{align}\label{eq:shat}
        s(\theta\mid D)\approx \hat s(\theta\mid D)=1-F_{\chi^2_d}(n\,\overline g(\theta)^\intercal\Sigma_G(\theta^*)^{-1}\overline g(\theta))
    \end{align}
\end{theorem}
\begin{proof}
    The proof follows from Lem. \ref{lem:credible} by rewriting the quadratic term of Lem. \ref{lem:credible} using Lem. \ref{lem:inv} as $(\theta-\theta^*)^\intercal H(\theta-\theta^*)=n^2\,\overline g(\theta)^\intercal H^{-1}HH^{-1}\overline g(\theta)=n^2\,\overline g(\theta)^\intercal H^{-1}\overline g(\theta)\approx n\,\overline g(\theta)^\intercal\Sigma_G^{-1}\overline g(\theta)$, where in the last step we have used Lem.  \ref{lem:infeq3} to obtain $H^{-1}\approx  \Sigma_G^{-1}/n$.
\end{proof}

We prove an auxiliary theorem which allows us to obtain an uncertainty estimate for any scalar function by using the gradient covariance. In practical computations, we can use any value of $\theta$ near the MAP solution $\theta^*$ as $\Sigma_G(\theta)$ and $f'(\theta)$ do not typically vary much near $\theta^*$.
\begin{theorem}[Uncertainty of a Scalar Function via Gradient Covariance]\label{thm:std}
Let $f(\theta)$ be a differentiable affine function with gradient \( f'(\theta)=\nabla_\theta{f(\theta)} \). \(\theta\) is drawn from the quadratic posterior (Ass. \ref{ass:gauss}) with $E[\theta] = \theta^*$. The first-order Taylor expansion about \(\theta^*\) yields that the standard deviation of \(f(\theta)\) is given by
$\sigma[f(\theta)]\approx \sqrt{f'(\theta^*)^\intercal \Sigma_G(\theta^*)^{-1} f'(\theta^*)/n}$. 
\end{theorem}
\begin{proof}
Using a first-order Taylor expansion of \(f\) around the posterior mean \(\theta^*\):
\begin{equation}
f(\theta) = f(\theta^*) +  (\theta-\theta^*)^\intercal f'(\theta^*)\,.
\end{equation}
We have $E[f(\theta)] = f(E[\theta])$ as $f$ is affine and
hence we have $f(\theta)-E[f(\theta)] =  (\theta-\theta^*)^\intercal f'(\theta^*)$. The variance is, therefore, given by 
\begin{align}
\operatorname{Var}\bigl[f(\theta)\bigr]&= E\left[\left(f(\theta)-E[f(\theta)]\right)^\intercal \left(f(\theta)-E[f(\theta)\right)\right]\nonumber\\
&=
f'(\theta^*)^\intercal E\left[(\theta-\theta^*)(\theta-\theta^*)^\intercal\right] f'(\theta^*)\nonumber\\
&=f'(\theta^*)^\intercal H^{-1}f'(\theta^*),
\end{align}
where we have used Ass. \ref{ass:gauss} to get $E\left[(\theta-\theta^*)(\theta-\theta^*)^\intercal\right]=H^{-1}$.
We obtain the claim by taking the square root and using Lem.~\ref{lem:infeq3} to substitute \( H^{-1}\approx \Sigma_G(\theta^*)^{-1}/n \).
\end{proof}

 \section{Practical Considerations}\label{sec:practical_considerations}

 In this section, we outline some important considerations regarding the practical implementation of our method.

 \paragraph{Numerical Stability of Covariance Matrix Inversion}  
 The empirical gradient covariance matrix $\Sigma_G$ may not be of full rank and cannot be directly inverted, particularly when $n<d$, one of the primary scenarios where our method is intended to be used. To address this issue, we use Oracle Approximating Shrinkage (OAS) to estimate the covariance matrix, resulting to full-rank estimate $\hat \Sigma_G=(1-\epsilon)\Sigma_G+\epsilon\,{\rm tr}(\Sigma_G){\bf 1}_d/d$, where $\epsilon\in[0,1]$ is the OAS regulariser computed as in \cite{chen2010Shrinkage}.

 \paragraph{Computational Complexity of Covariance Matrix Inversion}
 The quadratic term in Thm. \ref{thm:main} $n\,\Bar{g}(\theta)^{\top}\hat\Sigma_G^{-1}\Bar{g}(\theta)$ can be computed in $O((n+d)d^2)$ time. However, if $n<d$, which is the case in our experiments, we can use the Woodbury matrix identity, which results in the same numerical value but can be computed in $O((n+d)n^2)$ (lines 8--9 of Alg. \ref{alg:credible}).

 \paragraph{Scaling of the Loss Function} 
Our derivations assume that the posterior is of the form $p(\theta\mid D)\propto e^{-L(\theta)}$. Multiplying the loss by a constant factor $\kappa\in{\mathbb{R}}_{>0}$ $L(\theta)\rightarrow\kappa\,L(\theta)$ corresponds to scaling the likelihoods as $p(z\mid\theta)\rightarrow p(z\mid\theta)^\kappa/Z_\kappa$, prior as   $p(\theta)\rightarrow p(\theta)^\kappa/Z^0_\kappa$, the quadratic form of Thm. \ref{thm:main} as $n\,\Bar{g}(\theta)^{\top}\tilde\Sigma_G^{-1}\Bar{g}(\theta)\rightarrow\kappa\,n\,\Bar{g}(\theta)^{\top}\tilde\Sigma_G^{-1}\Bar{g}(\theta)$, and the uncertainty estimate of Thm. \ref{thm:std} as $\sqrt{f'(\theta^*)^\intercal \Sigma_G^{-1} f'(\theta^*)/n}\rightarrow \sqrt{f'(\theta^*)^\intercal \Sigma_G^{-1} f'(\theta^*)/(\kappa\,n)}$. In practical applications (where the loss function may have arbitrary scaling), we may need to rescale the loss. In our experiments, we have used unscaled loss and $\kappa=1$.

 \paragraph{Evaluating the Gradient Covariance}
 $\Sigma_G(\theta)$ does not depend on $\theta$ (hence, $\Sigma_G(\theta)=\Sigma_G(\theta^*)$) if and only if the Hessian $h(z,\theta)=\nabla_\theta^2{l(z,\theta)}$ is constant in $z$ and $\theta$, meaning that $l(z,\theta)=(\theta-\theta^*_z)^\intercal A (\theta-\theta^*_z)/2+c_z$ for some constant symmetric positive-definite (for the loss to have a unique minimum) matrix $A$ and vector $\theta^*_z$ and scalar $c_z$ (the latter being functions of only $z$), which is not exactly satisfied for most practical loss functions in machine learning.
  Even so, in our examples, $\Sigma_G(\theta)$ does not vary much with respect $\theta$ in practice, hence, $\Sigma_G(\theta)\approx \Sigma_G(\theta^*)$ can be used when applying Thms. \ref{thm:main} and \ref{thm:std}.

 \paragraph{Deterministic Stopping Condition}
In this paper, we propose a randomized stopping condition. However, it can be easily modified to a deterministic one by replacing the random draw (line~1 of Alg.~\ref{alg:gradstop}) with a constant $u \in [0,1]$ specified by the user. In Sec.~\ref{sec:results}, we show that using a constant threshold value results in good performance in practice.

\section{Experimental Results}\label{sec:results}

In this section, we present the experimental results demonstrating the competitive performance of the proposed method {\sc gradstop} applied to tasks with limited access to training data. We consider
classification of tabular datasets arising in medical applications and classification of images in a transfer learning scenario. 

Our goal is to find the optimal stopping point for the optimisation of model parameters that maximizes model performance on an unseen test set. The test sets are reserved as 20\% from each dataset.
In each experiment, we set a computational budget of $N$ iterations and calculate the credibility value statistic (see Alg.~\ref{alg:credible}) and the statistics proposed in prior works on each iteration $t$.

We compare the results obtained by {\sc gradstop} with training for the full pre-defined computational budget and stopping of the training based on the validation set performance as well as the following baselines: Gradient Disparity (\textbf{GD}) \cite{forouzesh2021disparity}, Evidence bound (\textbf{EB}) \cite{mahsereci2017early}, Gradient signal to noise ratio (\textbf{GSNR}) \cite{liu2020understanding}, and the criteria based on gradient vector similarity calculated by applying $\sign$ and $\cos$ functions to the gradient inner product \cite{fort2019stiffness,forouzesh2021disparity}. 
For the runs with the validation set stopping criteria, we hold out 20\% of the medical datasets or the transfer set as the validation set.
GD baseline stops the training when the statistic value increases for the fifth time from the beginning of training. Similarly, GSNR, $\sign$ and $\cos$ baselines stop when the value of the monitored statistic decreases for the fifth time. EB proposes the stopping point when the value of the statistic becomes positive.
Since all the baselines provide a deterministic stopping criteria we use a deterministic version of our method for fair comparison. We stop the training with {\sc gradstop} when the credible value is above a threshold of $0.1$ in all of the experiments. We report the ablation result for the threshold value in Sect.~\ref{sec:ablations}. 
In all experiments, the model parameters are optimised with Adam \cite{kingma2014adam}. For all methods, we report mean and standard deviation of the loss and / or accuracy calculated on the test set across 10 random seeds.

Unlike the baselines, the proposed method allows estimating the uncertainty of the parameter values. We also present the provided estimates compared to MCMC in Sect.~\ref{sec:gradstop_vs_mcmc}.

\subsection{Training from scratch with limited data}
\label{sec:med_data}

Medical datasets are typically costly and time-consuming to collect, thereby using the full dataset for training is a highly desirable property. Moreover, even a small improvement in prediction accuracy implies more patients being diagnosed and treated correctly, further motivating the method development.

In this experiment, we consider three medical datasets describing the following conditions: 
heart disease \cite{heart_disease_45}, diabetes \cite{diabetes_34}, and hepatitis \cite{hepatitis_46}. For all datasets, we applied standardization to input features.
We train a classical logistic regression (LR) and a multi-layer perceptron (MLP) for this task. In case of LR the number of model parameters is lower or comparable to the number of available data points while MLP demonstrates the method performance in an overparametrised regime. Here, the MLP consists of three layers with 64, 128, and 32 neurons and an output layer with ReLU non-linearity and batch normalization \cite{ioffe2015batch} between the layers and softmax applied to the output. We ran LR and MLP models for the computational budget of 4000 and 300 epochs correspondingly.

The proposed method performs on par with the baselines both for LR and MLP models. Our method shows superior performance for diabetes and hepatitis datasets with MLP as a model and is among the top performing models with LR. For heart disease dataset, {\sc gradstop} outperforms the baselines with LR model (see Table~\ref{tab:med_ml}).

\begin{table}[h]
    \centering
    \caption{\textbf{Classification of medical data.} Test losses (and standard deviations) of the proposed  method {\sc gradstop} and prior early stopping techniques for classification of medical data with LR and MLP models.}
    \begin{tabular}{c|c|c|c}  %
        \hline
        \multirow{2}{*}{\textbf{Dataset}} & \multirow{2}{*}{\textbf{Method}} & \multicolumn{2}{c}{\textbf{Model}} \\
        \cline{3-4}
         & & \textit{LR} & \textit{MLP} \\
        \hline
        \multirow{8}{*}{\textit{Heart Disease}} & End of training & 1.468 (0.031)& 2.182 (0.183) \\
         & Validation set & 1.138 (0.128)& \textbf{1.257} (0.133) \\
         & GD & 1.828 (0.431)& 1.354 (0.201) \\
         & EB & 1.096 (0.099)& 2.182 (0.183) \\
         & GSNR & 1.856 (0.414)& 1.386 (0.227) \\
         & $\sign$ & 1.856 (0.414)& 1.398 (0.206) \\
         & $\cos$ & 1.856 (0.414)& 1.389 (0.236) \\
         & {\sc gradstop} (ours) & \textbf{1.055} (0.082)& 1.52 (0.308) \\
        \hline
        
        \multirow{8}{*}{\textit{Diabetes}} & End of training & \textbf{0.511} (5e-07)& 1.988 (0.183) \\
         & Validation set & 0.511 (0.003)& 0.907 (0.175) \\
         & GD & 0.888 (0.305)& 0.953 (0.094) \\
         & EB & 0.522 (0.024)& 1.988 (0.183) \\
         & GSNR & 0.902 (0.319)& 0.944 (0.110) \\
         & $\sign$ & \textbf{0.511} (5e-07)& 1.027 (0.088) \\
         & $\cos$ & 0.561 (0.108)& 0.925 (0.116) \\
         & {\sc gradstop} (ours) & 0.518 (0.018)& \textbf{0.785} (0.155) \\
        \hline

        \multirow{8}{*}{\textit{Hepatitis}} & End of training & 0.46 (8e-05)& 2.758 (0.756) \\
         & Validation set & \textbf{0.403} (0.151)& 0.657 (0.199) \\
         & GD & 0.963 (0.404)& 1.026 (0.292) \\
         & EB & 0.481 (0.125)& 2.758 (0.756) \\
         & GSNR & 0.963 (0.404)& 0.98 (0.270) \\
         & $\sign$ & 0.46 (8e-05)& 1.312 (0.353) \\
         & $\cos$ & 0.465 (0.149)& 1.028 (0.270) \\
         & {\sc gradstop} (ours) & 0.415 (0.085)& \textbf{0.625} (0.141) \\
        \hline

    \end{tabular}
    \label{tab:med_ml}
\end{table}

\subsection{Transfer learning for image classification}

\begin{figure}
    \centering
    \vspace{0.3cm}
    \includegraphics[width=\linewidth]{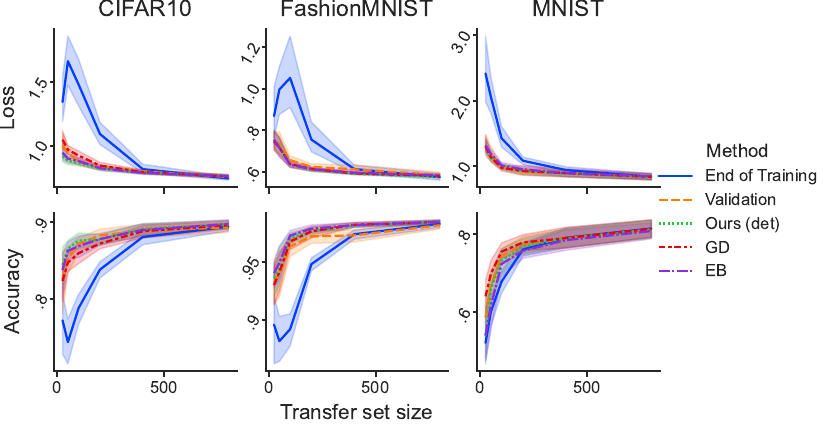}
    \vspace{-0.5cm}
    \caption{\textbf{Method evaluation for transfer learning with image data.} Comparison between the test losses at the end of transfer learning (blue), at the minimum of validation loss (orange) and at stopping points found by our method (green), and by closest baselines GD (red) and EB (purple). All of the early stopping methods outperform running the transfer learning for the full budget (blue), and our method slightly outperforms the baselines for CIFAR10 dataset.}
    \label{fig:transfer_vs_validation}
\end{figure}

\begin{figure}[h]
    \centering
    \includegraphics[width=\linewidth]{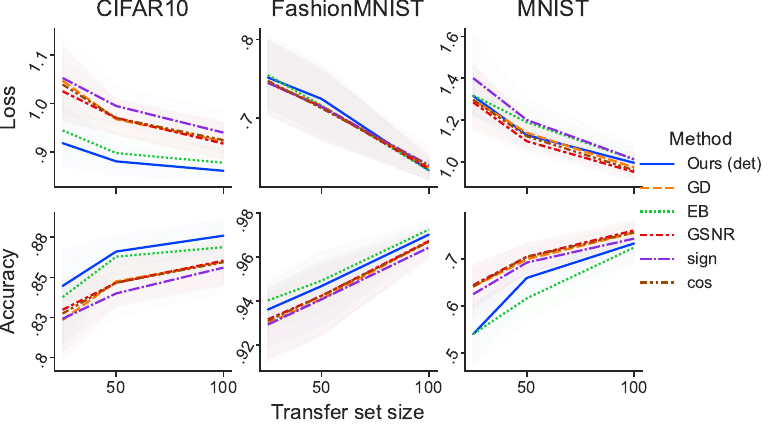}
    \vspace{-0.5cm}
    \caption{\textbf{Method comparison with an extended list of baselines for transfer learning.} Test losses and accuracy for early stopping of transfer learning task with {\sc gradstop} (solid blue line) and baseline methods: GD (orange), EB (green), GSNR (red), $\sign$ (purple), $\cos$ (brown).}
    \label{fig:baselines_zoomed}
\end{figure}

We further demonstrate the effectiveness of the proposed method for identifying a suitable model checkpoint for transfer learning tasks with varying transfer set sizes.
We run the experiment on three well-known image classification datasets: MNIST \cite{lecun1998gradient}, FashionMNIST \cite{xiao2017fashion}, and CIFAR-10 \cite{krizhevsky2009learning}.
In this experiment, we reserve a subset of classes for transfer learning while the images belonging to the remaining classes are used for model pre-training. For MNIST dataset, digits from zero to four are included in the model pre-training while the transfer learning task is to correctly classify images of digits from five to nine. Similarly, the experiments with FashionMNIST and CIFAR-10 data have two holdout classes unseen in pre-training and used solely for transfer learning.
For this task, we train a neural network with three convolutional layers with max-pooling followed by two fully-connected layers with ReLU activations between the layers and a softmax applied to the network output. In the transfer learning regime, only the output layer parameters are optimised while the rest of the model parameters are frozen.

Fig.~\ref{fig:transfer_vs_validation} shows that the proposed method, the closest baselines, such as GD and EB, and the validation set-based method provide suitable stopping criteria for the optimisation, whereas running transfer learning for the full computational budget of 1000 epochs results in overfitting.
The overfitting is observed for all studied datasets, especially with small transfer set sizes, indicating the importance of early stopping for better generalisation performance. 
Both {\sc gradstop} and baseline methods yield comparable results in terms of the loss function values and model accuracy on the test set.
Additionally, the methods are comparable in terms of computational and memory requirements.
All methods have to store an additional copy of the model weights corresponding to the best value of the monitored statistics.
The main difference between the methods lies in the calculation of the statistics.
The validation set-based method relies on a forward pass through the model with the validation set data and calculates the loss value. At the same time, {\sc gradstop} and other baseline methods without a validation set reuse the gradients calculated on the training set by backdrop to obtain their statistics. 

To closer compare the baselines we present a zoomed version of the results for stopping criteria without validation set in Fig.~\ref{fig:baselines_zoomed}. The proposed method {\sc gradstop} slightly outperforms the baselines for CIFAR-10 dataset and is close to the best performers on MNIST. All studied methods provide very close results for FashionMNIST and clearly prevent overfitting.

\subsection{Early stopping for data with noisy labels}

We further test the stability of the early stopping criteria in case the transfer set in the previous experiment is contaminated with varying percentage of mislabeled examples. We vary the ratio of noisy labels from 0.1 to 0.35 with 0.05 step and set the transfer size $n=100$.

\begin{figure}
    \centering
    \includegraphics[width=\linewidth]{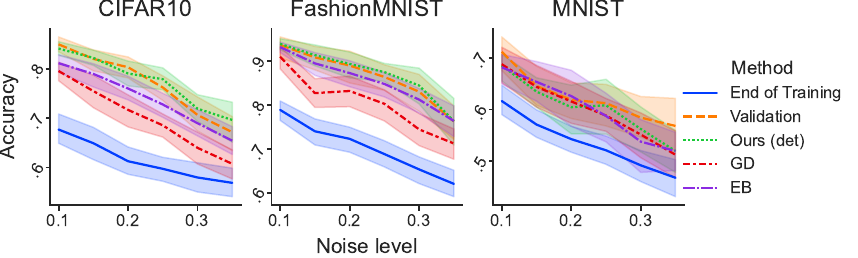}
    \vspace{-0.5cm}
    \caption{\textbf{Evaluation of robustness to noise in labels of the transfer set.} Model accuracy for transfer learning with transfer size $n=100$ for CIFAR, FashionMNIST, and MNIST datasets (from left to right) with varying level of noise in the labels of the transfer set. The early stopping methods, including {\sc gradstop}, are effective in the presence of noise compared to training for the full computational budget.}
    \label{fig:transfer_noisy}
\end{figure}

Fig.~\ref{fig:transfer_noisy} shows that the overfitting problem in training  with full computational budget still persists even in case of noisy labels, which provide implicit regularisation to the model training. All considered stopping criteria yield better generalisation results than a pre-defined computation budget, confirming that the finding from previous experiment hold in case of presence of noise in the labels.

\subsection{Uncertainty quantification}
\label{sec:gradstop_vs_mcmc}

\begin{figure}
    \centering\includegraphics[width=\linewidth]{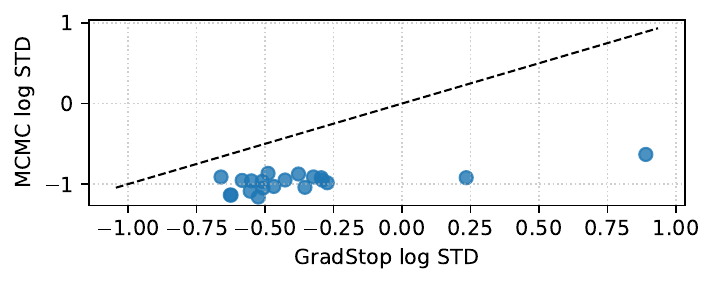}
    \vspace{-0.5cm}
    \caption{\textbf{Log standard deviation of model parameters.} Comparison of log standard deviation by {\sc gradstop} and MCMC for hepatitis dataset with logistic regression.}
    \label{fig:gradstop_vs_mcmc}
\end{figure}

In this experiment we demonstrate the uncertainty estimates of parameter values according to Thm.~\ref{thm:std}, which we compared with full Bayesian sampling by MCMC. Both methods fit a logistic regression model for a binary classification task of hepatitis cases \cite{hepatitis_46} (as in Sect.~\ref{sec:med_data}). The estimates for standard deviation by our method provide an upper bound on the parameters uncertainty (see Fig.~\ref{fig:gradstop_vs_mcmc}). The estimates can serve as a fast and conservative alternative for uncertainty quantification without the high computational cost associated with MCMC.

\subsection{Method ablations}
\label{sec:ablations}

\begin{figure}
    \centering
    \includegraphics[width=\linewidth]{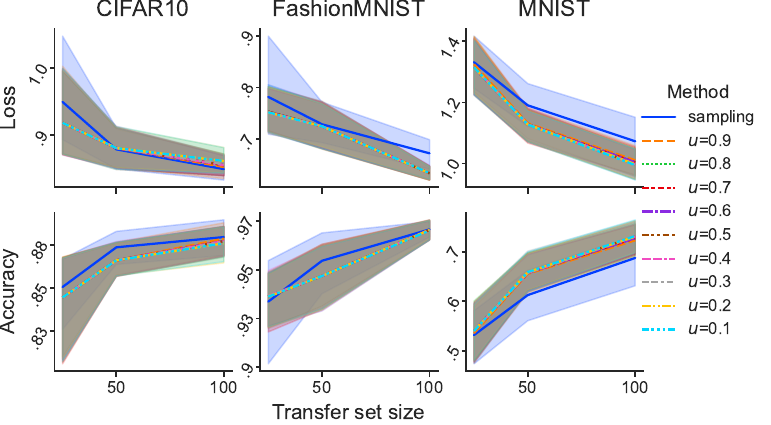}
    \vspace{-0.5cm}
    \caption{\textbf{Sensitivity to the value of threshold $u$.} Test set loss and accuracy for transfer learning experiments with a stochastic and a deterministic stopping with varying threshold $u$ in  {\sc gradstop}. The performance is stable under the changes in the hyperparameter value.}  \label{fig:transfer_oas_det_vs_sampling}
\end{figure}

We compare the effect of the selected threshold value $u$ for the deterministic version of {\sc gradstop} by varying the threshold value from 0.1 to 0.9. Fig.~\ref{fig:transfer_oas_det_vs_sampling} demonstrates that the reported results are robust to changes in the hyperparameter value, in particular if the value is selected below 0.5. Additionally, the deterministic stopping criteria typically results in more stable results with smaller standard deviation, especially for FashionMNIST and MNIST datasets.

\section{Future Work}\label{sec:future_work}

This work can be further extended in several ways.
The current version of {\sc gradstop} relies on the availability of the full-batch gradients. To make the algorithm more widely applicable, future work could explore extensions to stochastic gradient descent \cite{robbins1951stochastic} and other optimisation settings without full-batch gradient information \cite{ruder2016overview}. A straightforward implementation would be to compute the stopping statistics with gradients of one batch for each training epoch.

Most machine learning models are not fully probabilistic and rely on various loss functions for training \cite{goodfellow2016deep,terven2023loss}. In this paper, the models we considered are not inherently probabilistic; still, they can all be decomposed into a sum of datapoint-specific terms as in Eq. \eqref{eq:losssum}. A promising direction for future research is to extend our approach to accommodate a broader range of loss functions. This could enable the joint training of several models, such as a generator and a discriminator \cite{goodfellow2014generative},  or optimising for multiple objectives \cite{sener2018multi,kurin2022defense}.

More broadly, it is known that many deep learning models work and produce astonishing results in over-parametrised regimes; thus, many lessons learnt from statistics and classical machine learning do not hold. The optimisation of neural network weights using stochastic gradient descent exhibits phenomena of double-descent \cite{belkin2019reconciling,nakkiran2021deep,heckel2021early} (the validation loss decreases initially, then it starts increasing and then decreases again to even lower values than was achieved before in the ``first descent'') and grokking (generalisation occurs well after the training loss goes near zero) \cite{power2022grokking,humayun24deep}. These phenomena require better theoretical understanding and practical techniques to find a suitable stopping time for training in order to achieve optimal predictive performance.

\section{Conclusion}\label{sec:conclusion}

In this paper, we introduced {\sc gradstop}, a novel early-stopping stochastic method based on posterior sampling designed for scenarios where only gradient information is available.
Unlike conventional early stopping techniques that rely on a separate validation set, our method approximates the posterior distribution along the gradient descent trajectory. It determines an optimal stopping point by sampling from the level sets of the approximated posterior distribution.

By framing early stopping as a posterior sampling problem, we establish a connection between Bayesian inference and optimisation-based training procedures, providing a principled stopping criterion that is theoretically grounded.
In the theoretical section, we demonstrated that if the loss near its minimum can be approximated by a quadratic function, 
gradient information is sufficient for sampling from the level sets of the posterior distribution.
Furthermore, our empirical experiments show that {\sc gradstop} effectively determines stopping points that yield strong generalisation performance. 

The proposed algorithm provides suitable stopping criteria for gradient-based optimisation. Our method performs comparably to baselines across various tasks while utilising all available training data and providing uncertainty estimates for parameters, making it particularly beneficial in data-constrained settings. The evaluation of the introduced statistics adds a small computational overhead, however, the statistic could be integrated with gradient descent optimisation libraries for practical use.

\ack We thank the Research Council of Finland grants 364226 and 345704 related to the Virtual Laboratory for Molecular Level Atmospheric Transformations (VILMA) Centre of Excellence, and Flagship Programme: Finnish Center for Artificial Intelligence (FCAI), and the Helsinki Institute for Information Technology HIIT for funding. We thank the Finnish Computing Competence Infrastructure (FCCI) for providing computational resources.

\bibliography{bibliography}

\end{document}